\documentclass[%
 reprint,
superscriptaddress,
 amsmath,amssymb,
 aps,
pra,
]{revtex4-2}

\usepackage{graphicx}
\graphicspath{{./figs/}}
\usepackage{caption}

\usepackage{subcaption}
\usepackage{dcolumn}
\usepackage{bm}
\usepackage{algorithm,algcompatible}
\usepackage[usenames,dvipsnames]{color} 
\usepackage{ifthen} 
\usepackage{braket}
\usepackage{amsmath}
\usepackage{array}
\usepackage{tikz}
\newcommand*\circled[1]{\tikz[baseline=(char.base)]{
            \node[shape=circle,draw,inner sep=2pt] (char) {#1};}}
\usepackage{mathtools}
\DeclarePairedDelimiter{\abs}{\lvert}{\rvert}
\usepackage[toc,page]{appendix}

\def\shownoteal{1} 
\newcommand{\nal}[1]{\ifthenelse{\shownoteal=1}{\textcolor{red}{[[#1]]}}{}}

\begin{document}

\preprint{APS/123-QED}

\title{QubitHD: A Stochastic Acceleration Method\\ for HD Computing-Based Machine Learning}

 \author{Samuel Bosch}
 \email{sbosch@mit.edu}
 \address{
 Massachusetts Institute of Technology, Cambridge, MA 02139, USA\\
}

\author{Alexander Sanchez de la Cerda}
\address{
 Harvard University, Cambridge, MA 02138, USA\\
}

\author{Mohsen Imani}
\address{
University of California Irvine, Irvine, CA 92697, USA\\
}

\author{Tajana~\v{S}imuni\'{c} Rosing}%
\address{
University of California San Diego, La Jolla, CA 92093, USA\\
}

\author{Giovanni~De Micheli}
\address{
\'{E}cole polytechnique f\'{e}d\'{e}rale de Lausanne, Lausanne, VD 1015, Switzerland\\
}

\date{\today}

\begin{abstract}
 
Machine Learning algorithms based on Brain-inspired Hyperdimensional (HD) computing imitate cognition by exploiting statistical properties of high-dimensional vector spaces. It is a promising solution for achieving high energy efficiency in different machine learning tasks, such as classification, semi-supervised learning, and clustering. A weakness of existing HD computing-based ML algorithms is the fact that they have to be binarized to achieve very high energy efficiency. At the same time, binarized models reach lower classification accuracies. To solve the problem of the trade-off between energy efficiency and classification accuracy, we propose the \textit{QubitHD} algorithm. It stochastically binarizes HD-based algorithms, while maintaining comparable classification accuracies to their non-binarized counterparts. The FPGA implementation of \textit{QubitHD} provides a 65\% improvement in terms of energy efficiency, and a 95\% improvement in terms of training time, as compared to state-of-the-art HD-based ML algorithms. It also outperforms state-of-the-art low-cost classifiers (such as Binarized Neural Networks) in terms of speed and energy efficiency by an order of magnitude during training and inference.

\end{abstract}


\maketitle


\section{\label{sec:introduction}Introduction:\protect
}
With the rise of data science and the Internet of Things (IoT), the amount of produced data on a daily basis has increased to a level we are barely able to handle \cite{gubbi2013internet}. As the amount of data that needs to be processed is often significantly larger than small-scale and battery-powered devices can handle, so many of these devices are forced to connect to the internet to process the data in the cloud. 
Deep Neural Networks (DNNs) are used for complex classification tasks, such as text and image recognition \cite{deng2009imagenet}, translation, and even medical applications \cite{mu2022augmenting}. However, the complexity of DNNs makes them impractical for some real-world applications, such as classification tasks on small battery-powered devices. Engineers often face a trade-off between energy efficiency and the achieved classification accuracy. Therefore, we need to create lightweight classifiers, which can perform inference on small-scale operating devices.\\ 

Brain-inspired Hyperdimensional (HD) computing \cite{kanerva2009hyperdimensional} has been proposed as a lightweight learning algorithm and methodology. The principles governing HD computing are based on the fact that the brain computes with \emph{patterns of neural activities} which are not directly associated with numbers \cite{kanerva2009hyperdimensional}. Machine learning algorithms based on Brain-inspired HD computing imitate cognition by exploiting statistical properties of very high-dimensional vector spaces. Recently, architectures such as \textit{LookHD} \cite{imani2021revisiting} have been proposed, which enable real-time HDC learning on low-power edge devices. The first step in HD computing is to map each data point into a high-dimensional space (e.g., $10,000$ dimensions). During training, HD computing linearly combines the encoded hypervectors to create a hypervector representing each class. During the inference, classification is done by calculating the cosine similarity between the encoded query hypervector and all class hypervectors. The algorithm then predicts the class with the highest similarity score. In the case of multiple classes with high similarity, the algorithm is likewise suited to express confidence in the correctness of a prediction.

Many publications on Brain-inspired HD computing argue that for most practical applications, HD computing has to be trained and tested using floating-point, or at least integer values \cite{morris2019comphd, imani2019sparsehd}. Binarized HD computing models provided low classification accuracies. Often too low for practical applications. An algorithm called \textit{QuantHD} \cite{QuantHD} revealed the existence of a method to improve the classification accuracies of binarized and ternarized models significantly. Nevertheless, there still exists a large gap between the classification accuracy of non-binarized and binarized HD computing classifiers. Also, such methods increase the required training time and are unstable as they tend to get stuck in local minima during training. In this paper, we propose a new method that can, both, reduce this classification accuracy gap by between a third and a half whilst simultaneously improving energy efficiency during training by \textbf{60\%}, on average. It also makes the training more stable by introducing randomness. We call this technique \textit{QubitHD}, as it is based on the principle of information being stored in a quantum bit (Qubit) before its measurement. The floating-point values represent the quantum state, while the binarized values represent the quantum state after a measurement has been performed. \\

The main contributions of the paper are the following:
\begin{itemize}
    \item We decreased the gap in classification accuracy between binarized and non-binarized state-of-the-art HD computing-based ML algorithms by \textbf{38.8\%}, on average.
    \item We decrease the convergence time in the range of \textbf{30-50\%} (for different datasets). Introducing randomness in the algorithm prevents it from getting stuck in local minima, and incites the algorithm to quickly move towards the optimal value. The reason why the authors of \cite{QuantHD} had problems with slow convergence was precisely this: lack of randomness.
    \item \textit{QubitHD} performs a similarity check by calculating the Hamming distance between the hypervectors instead of calculating the costly cosine similarity.
    \item We implemented the algorithm on FPGA, which accelerates training and inference. We also evaluated several classification problems, including human activity, face, and text recognition. When looking at energy efficiency and speed, the FPGA implementation of \textit{QubitHD} provides, on average, a \textbf{56}$\times$ and \textbf{8}$\times$ energy efficiency improvement and speedup during training, as compared to state-of-the-art HD computing algorithms \cite{ISLPED}. For comparison purposes, the authors of \cite{QuantHD} only achieve 34.1$\times$ and 4.1$\times$ energy efficiency improvement and speedup during the training against the same state-of-the-art HD computing algorithms.
    When comparing \textit{QubitHD} with multi-layer perceptron (MLP) and binarized neural network (BNN) classifiers, we observe that \textit{QubitHD} can provide \textbf{56}$\times$ and \textbf{52}$\times$ faster computing in training and testing respectively, while providing similar classification accuracies (see Table \ref{tab:compare}).
\end{itemize}

\begin{figure} [h]
\centerline{
{\includegraphics[width=1\columnwidth]{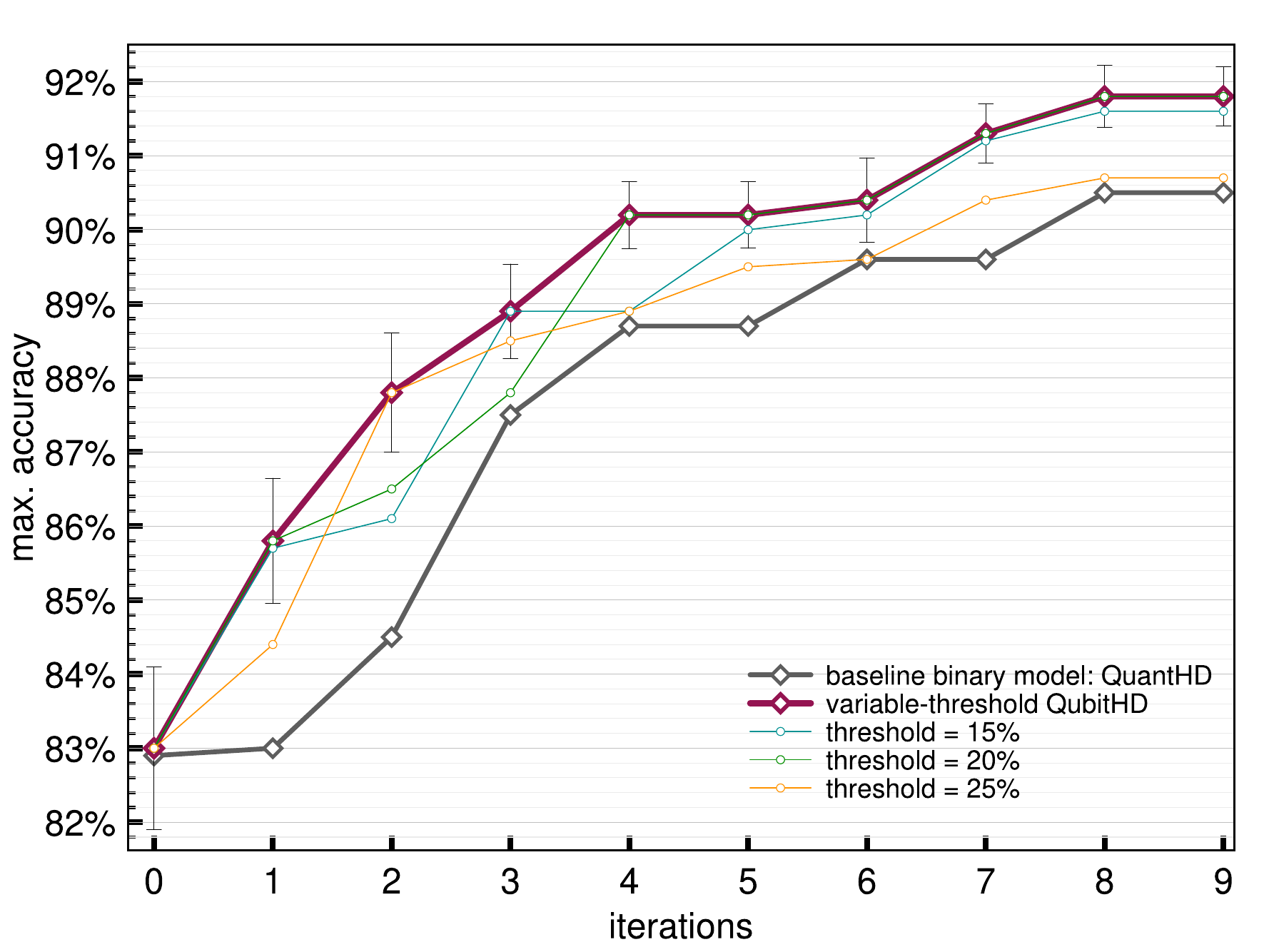}}
}
\caption{Average classification accuracies during different retraining iterations of \textit{QubitHD} compared to binary \textit{QuantHD}, for datasets listed in table (\ref{tab:accuracy}). It is clear that \textit{QubitHD} converges faster than \textit{QuantHD}, as a result of the stochastic binarization process. It does so without utilizing additional hardware resources. Table (\ref{tab:accuracy}) also lists the maximum accuracies achieved with a large number ($\geq 1000$) of training iterations for the individual datasets. 
}
\label{fig:accuracyy}
\end{figure}

\section{Hyperdimensional Computing}
\begin{figure*}[htp]
\centerline{
\includegraphics[width=0.8\textwidth]{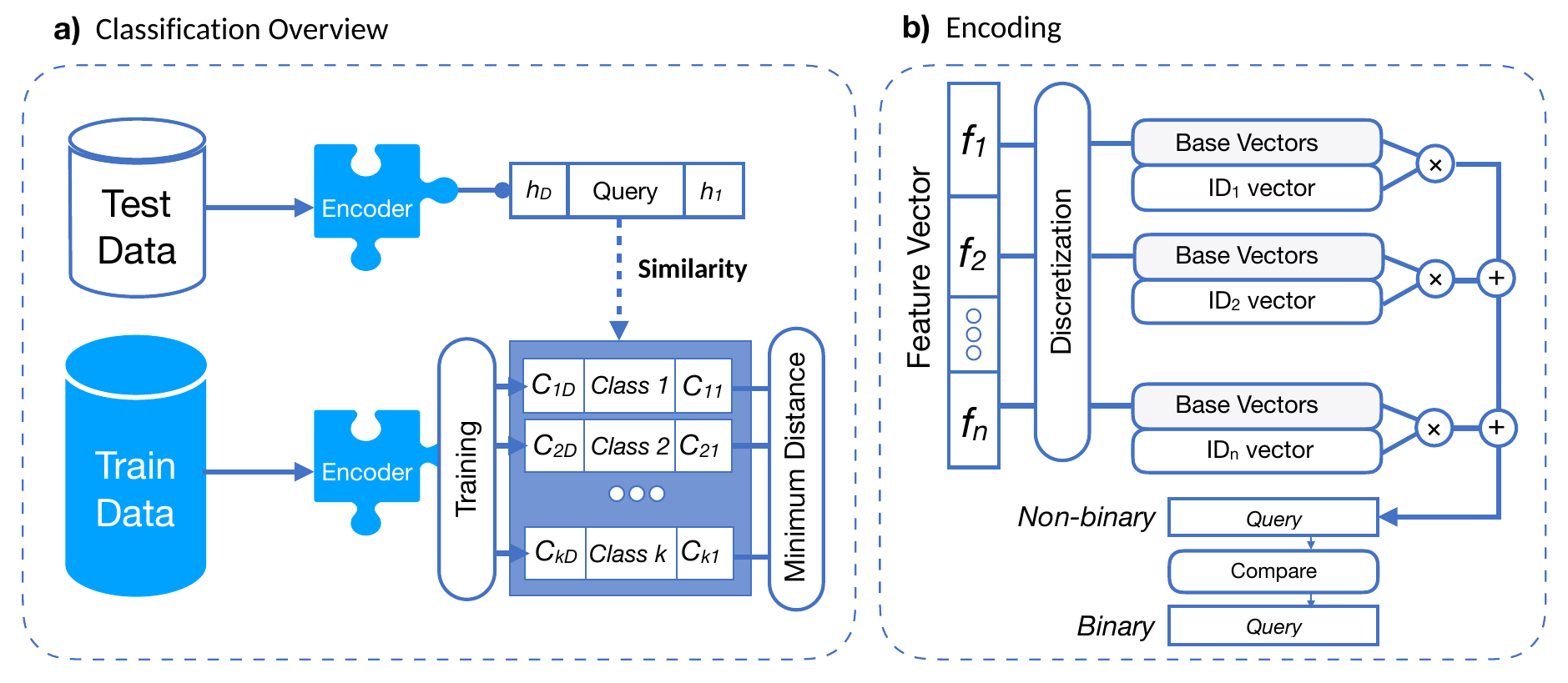}
}
\caption{Overview of a simple HD computing-based machine learning algorithm for classification on the left. The encoding scheme used in this publication is illustrated on the right.}
\label{fig:HD}
\end{figure*}

The applications of brain-inspired HD computing in machine learning are diverse. In this publication, we only focus on supervised classification tasks, but a recent publication indicated that HD computing-based ML algorithms can be applied to clustering and semi-supervised learning as well \cite{SemiHD}. The basis of \textit{QubitHD} is described in Figure \ref{fig:HD}. The core difference to \textit{QuantHD} is the binarization step that is discussed in Section \ref{sec:stochastic_binarization}. The non-binarized algorithm with retraining consists of the following steps:

\subsection{\textbf{Encoding}}\label{sec:encoding} 

The training dataset is pre-processed by converting all data points into very high-dimensional vectors (hypervectors). We used hypervectors of length $D=10,000$ in this paper, as it is the standard baseline for all HD computing-based machine learning algorithms. Like explained in \cite{QuantHD}, the original data is assumed to have $n$ features: $\mathbf{f} = \{ f_{1}, \dots f_n \}$. The objective is encoding each feature that corresponds to each datapoint into the hypervector of dimension $D$ ($D=10,000$ in this paper). Each feature vector "memorizes" the value and position of the relevant feature. In order to take into account the position of each feature, we use a set of randomly generated base hypervectors $\{\mathbf{B}_i, \mathbf{B}_2, \dots, \mathbf{B}_n\}$, where $n$ is the total number of features in each data point ($\mathbf{B}_i \in \{-1,1\}^D$). Since the base-hypervectors are uniformly generated at random (with equal probability for $-1$ and $1$), they are approximately mutually orthogonal. The cosine product between hypervectors ranges between $\cos(H_1,H_2) \in \big[-1,1 \big]$. The expected cosine product of independent and randomly generated base-hypervectors is $\mathbb{E}[\cos(\textbf{B}_i,\textbf{B}_j)]=0$, whereas the variance is $V[\cos(\textbf{B}_i,\textbf{B}_j)]=\frac{1}{\sqrt{D}} \approx 0$ for $D >> n$ (random walk) for $i \neq j$. Thereby, the hypervectors are almost orthogonal. This is true only when the number of randomly generated base-hypervectors is significantly smaller than the dimension of the entire vector space $D$. For comparison of the binarized hypervectors, we use the Hamming distance. Therefore:
$$ \mathbb{E}[\delta(\mathbf{B}_i,~\mathbf{B}_j)]= D/2~~~~~~(for ~~i\neq j).$$
Here $\delta$ is the Hamming distance similarity between the two binarized base-hypervectors.

\subsection{\textbf{Initial training}} \label{sec:initial_training}
The first training round is performed by summing up all hypervectors pertaining to the same class. That is, we abstract all hypervectors with the same labels. This method is called one-shot learning and is, at the moment, the most widespread way of using HD computing in machine learning \cite{ISLPED, Mitra_HD_1, Mitra_HD_2, rahimi2016hyperdimensional, ISSCC}. We now have one matrix $C$ of size $m \times D$ ($C \in \mathbb{R}^{m \times D}$), where $m$ is the number of existing classes and $D$ is the length of the hypervectors. 

\subsection{\textbf{Retraining}} \label{sec:retraining}
The classification accuracy of the current model during the inference is low \cite{QuantHD}. For this reason, we have to do retraining. As displayed in Figure \ref{fig:framework}, retraining is done in the following way. We go through the entire dataset of encoded data points and test them to ascertain if they are correctly classified by the current model $C$. For every misclassified data point, we have to make additional improvements to the model. Let us assume that the correct label of a data point is $k$, but it was incorrectly classified as $l$. We now add the erroneously classified hypervector to its corresponding row $C_k$. (to make them more similar). We also subtract the incorrectly classified hypervector from the row corresponding to the inaccurately predicted class $C_l$ (to make them more distinct). To decrease the convergence of time and noise, it is common practice to introduce a learning rate of $\alpha$ in this step as illustrated in Figure \ref{fig:framework}a \cite{imani2019adapthd}. This process is repeated several times. 

\subsection{\textbf{Inference}} \label{sec:inference}
During the inference, we predict the class to which the data point belongs. This data point is encoded as described in \ref{sec:encoding}, and then compared to all the class hypervectors. The algorithm then predicts the class with the largest cosine similarity.

\subsection{\textbf{Binarization:}}
\begin{figure*} [t!]
\centerline{
\includegraphics[width=0.8\textwidth]{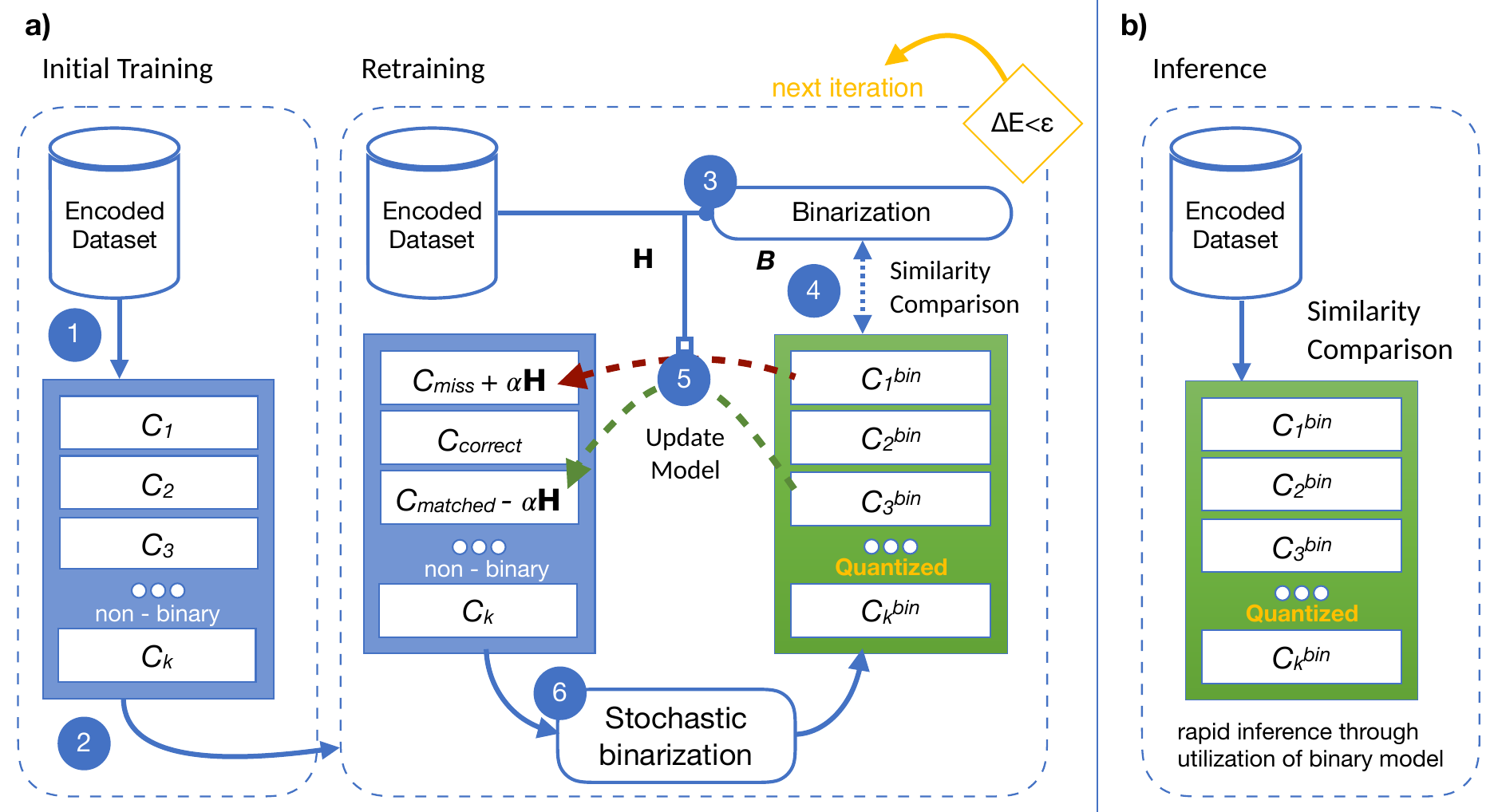}
}
\caption{(a) \textit{QubitHD} framework overview (including the initial training and the retraining of the non-binarized model based on the binarized model).  (b) Efficient inference from model}
\label{fig:framework}
\end{figure*}

So far, we described in the algorithm that the trained model has non-binarized elements. 
Many existing HD computing methods~\cite{rahimi2017high, imani2017exploring, rahimi2017hyperdimensional2} binarize the class hypervectors to eliminate costly \textit{cosine} operations used for the associative search ($C \in \mathbb{R}^{m \times D}  \rightarrow C^{binarized} \in \{-1,1 \}^{m \times D}$).   
Binary hypervectors do not provide sufficiently high classification accuracies on many (if not most) real-world applications. The usual way of binarizing class hypervectors is making all positive values equal to $+1$ and negative values equal to $-1$. This method suffers from a significant loss of information about the trained model. To the best of our knowledge, \cite{QuantHD} was the first publication demonstrating a method of achieving high classification accuracy while using a binarized (or quantized) HD model. Instead of just "blindly" binarizing the class hypervectors after every retraining iteration, \textit{QuantHD} trains the model in a way that is optimized for binarized hypervectors. That is, during every single retraining iteration, they create an additional binarized model. Doing so requires no additional computational power, as the binary representation of numbers in usual computer architectures reserves the first bit for the sign ($0$ stands for positive, $1$ for negative). The \textit{QuantHD} algorithm retrains on the predictions of the binarized model, while updating the non-binarized model as described in subsection \ref{sec:retraining} and Figure \ref{fig:framework}. The binarized model achieves, after several iterations, very high classification accuracies. They are significantly higher than they would be without binary-optimized retraining. \\

\section{Stochastic binarization} \label{sec:stochastic_binarization}
Our goal is to create a binarized model whose expected value is equal to the non-binarized model. This idea was inspired by the way in which quantum bits (or qubits) are measured – hence the name \textit{QubitHD}.\\

In other words, we want $\mathbb{E}\big[C^{\text{binarized}}\big] \approx C^{\text{non binarized}}$. The \textit{QuantHD} algorithm uses the following (very trivial) binarization function:
\begin{equation}
bin(x) = 
\begin{cases}
\label{eq:bin}
    1,& \text{if } x\geq 0\\
    -1,              & \text{otherwise}
\end{cases}
\end{equation}

We propose using the following method instead:

\begin{equation}
\label{eq:qbin}
qbin(x) = 
\begin{cases}
    1,& \text{if } x > b \\
    1,& \text{if } \abs{x} \leq b \text{, then with probability } \frac{1}{2}+\frac{x}{2b} \\
    -1,& \text{if } \abs{x} \leq b \text{, otherwise } \\
    -1,& \text{if } x < -b
\end{cases}
\end{equation}
where $b$ is the cutoff value defined as a fixed fraction of the standard deviation $\sigma$ of the data. It is discussed in greater detail in subsection \ref{sec:framework}. The advantage of doing so is the fact that the expected value of $qbin(x)$ for $x \in \big[-b,b\big]$ is proportional to x:
\begin{equation*}
    \mathbb{E}\big[ qbin(x) \big] =  (+1)(\frac{1}{2}+\frac{x}{2b})+(-1)(\frac{1}{2}-\frac{x}{2b})  = \frac{x}{b}
\end{equation*}

\section{Proposed \textit{QubitHD} algorithm}
\textbf{Motivation}

The \textit{QuantHD} algorithm still leaves us with a significant gap between the maximum classification accuracy of the floating-point model and the binarized one. Also, the \textit{QuantHD} retraining method described in \cite{QuantHD} tends to get "stuck" in local minima. Further, their algorithm almost doubles the average convergence rate, which hereafter increases energy consumption during training.\\
To summarize, here are the main problems with the \textit{QuantHD} algorithm, which the \textit{QubitHD} algorithm can either solve or improve:

\begin{enumerate}
\label{list:problems}
    \item There still exists a significant \textbf{gap} between the binarized model and non-binarized model accuracy
    \item The algorithm can sometimes get stuck in \textbf{local minima}, which makes it unreliable. 
    \item The \textbf{convergence time} of \textit{QuantHD} algorithm is almost twice as slow as compared to the other state-of-the-art HD computing algorithms with retraining
\end{enumerate} 
\medskip

\subsection{Framework of the \textit{QubitHD} algorithm} \label{sec:framework}
In this section, we present the \textit{QubitHD} algorithm. It enables efficient binarization of the HD model with a minor impact on classification accuracy. The algorithm is based on \textit{QuantHD} and consists of four main steps:

\textbf{1) Encoding}: This part is described in detail in Subsection \ref{sec:encoding} and Figure \ref{fig:HD}b\\

\textbf{2) Initial training}: \textit{QubitHD} trains the class hypervectors by summing all the encoded data points corresponding to the same class as seen in Figure \ref{fig:framework}a

It is evident from Figure \ref{fig:framework}a that every accumulated hypervector represents one class. As explained in \cite{QuantHD}, in an application with $k$ classes, the initially trained HD model contains $k$ non-binarized hypervectors $\{\mathbf{C}_1,\dots,\mathbf{C}_k\}$, where $\mathbf{C}_i \in \mathbb{N}^D$ $\circled{1}$.\\ 

\textbf{3) Stochastic binarization}:
\label{sec:qbin}
This part is the main change with respect to the \textit{QuantHD} algorithm. A given class hypervector is created by summing together random hypervectors $\mathbf{h}$ of the type $\mathbf{h} \in \{-1,1\}^D$. Every element $C_{ij}$ in a class hypervector $\mathbf{C_i}$ (of class i) is the product of a "random walk". In other words, its distribution follows a binomial distribution with a probability mass function (pmf):
\begin{equation}
    p(C_{ij}=k) = \binom{n}{k}p^k(1-p)^{n-k} = \binom{n}{k}\frac{1}{2^n}
\end{equation}
where $p=\frac{1}{2}$ (as we have equal probabilities for $h_j = -1$ and $h_j = +1$), n is the number of randomly summed hypervectors for class i, and k is a possible value $C_{ij}$ can take. Note that $C_{ij} \in \{-n,...,0,...,n\}$.

Assuming that the number of hypervectors corresponding to every class in the dataset is large enough, the \textit{normal distribution} is a good approximation for modeling the binomial distribution

\begin{figure} [htp]
\centerline{
\includegraphics[width=\columnwidth]{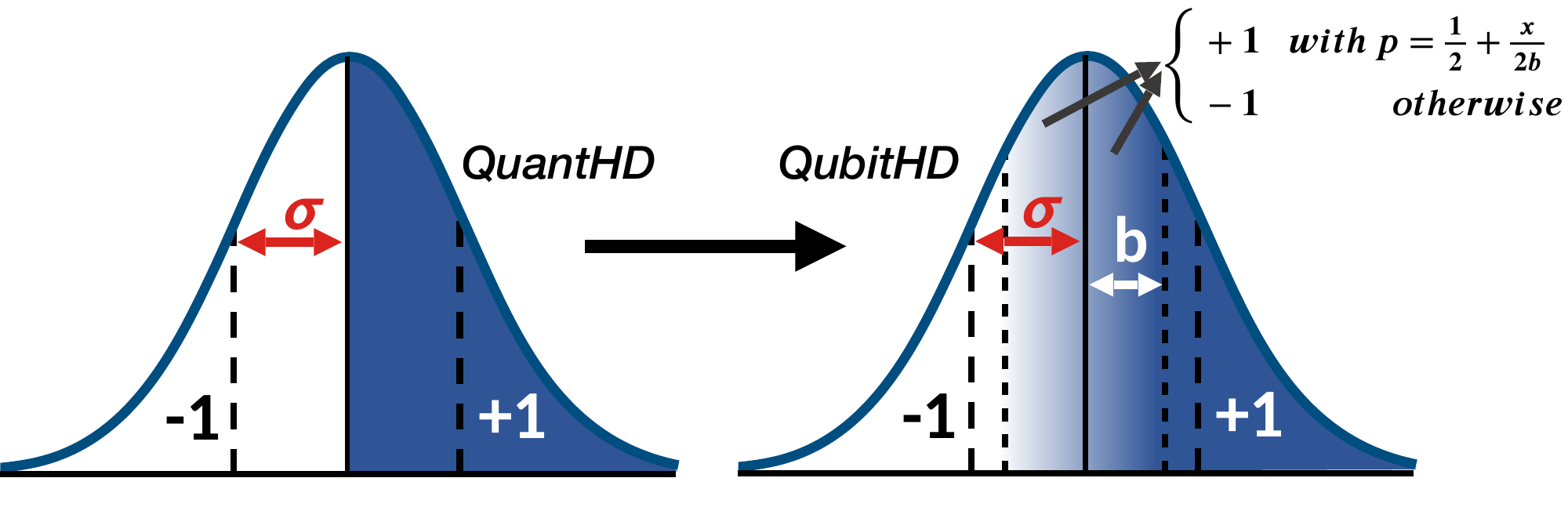}
}
\caption{On the left side, we have a visualization of equation \ref{eq:bin} from \textit{QuantHD}. On the right side, we have a visualization of equation \ref{eq:qbin} from \textit{QubitHD}. White represents $-1$, blue $+1$, and the colors between represent a stochastic selection.}
\label{fig:qbin}
\end{figure}

In previous publications, \cite{QuantHD}, the way of binarizing a model $C$ was described by Equation \ref{eq:bin}. We instead propose using Equation \ref{eq:qbin} shown in Figure \ref{fig:qbin}. Implementing this change requires almost no additional resources (the random flips have to be performed only once per retraining round), but leads to significant improvements in terms of accuracy, reliability, speed, and energy efficiency. The accuracy improvement is due to the fact that the expected value of this stochastically binarized model is equal to the non-binarized model. The reliability and speed improvement are because the model quickly "jumps" out of local minima, as opposed to getting stuck for several iterations. The energy consumption during training depends on the number of retraining iterations, which are significantly reduced. 

Just as we have demonstrated, the encoded data we are processing is (approximately) normally distributed. To be able to use Equation \ref{eq:qbin} for the binarization process, we have to define a "cutoff" value $b$. That is, everything above $+b$ will become $+1$, and everything below $-b$ will become $-1$. Only values between $-b$ and $+b$ will be better approximated through Equation \ref{eq:qbin}. In most cases, $b$ has to be smaller than the standard deviation of the data distribution $\sigma$. If we would use $b >> \sigma$, our model would become almost completely random, as most of the values are contained in the $\big[-\sigma,+\sigma\big]$ interval (68\% to be precise).

The reason why the $qbin(x)$ binarization works better than the $bin(x)$ lies in the fact that taking into account the expected value of the binarized model, it is equal to actual values in the non-binarized model, with the exception of values below $-b$ and above $+b$. Empirically, we also noticed that the randomness of $qbin(x)$ prevents the algorithm from getting stuck in local minima during training, which reduces the convergence time by \textbf{50\%}, on average. 

\section{Possible FPGA Implementation}~\label{sec:FPGA}
It is known that HD computing-based machine learning algorithms can be implemented in a wide range of different hardware platforms, such as CPUs, GPUs, and FPGAs. As most of the training and all of the inference relies on bit-wise operations, it was proposed in \cite{QuantHD} that FPGAs would be a suitable candidate for efficient hardware acceleration. 
The same hardware can be used for implementing both, the \textit{QuantHD} and the \textit{QubitHD} algorithm. This is also one of the major advantages of \textit{QubitHD}, as it doesn't require costly hardware upgrades from previous models.

\section{Evaluation}
\subsection{Experimental Setup}
The training and inference of the algorithm were implemented and verified using \textit{Verilog} and the code was synthesized on the \textit{Kintex-7 FPGA KC705 Evaluation Kit}. The \textit{Vivado XPower tool} has been used to estimate the device power. Additionally, for testing purposes, all parts of the \textit{QubitHD} algorithm have been implemented on the CPU. We also implemented the algorithm on an embedded device (\textit{Rasberry Pi 3}) with an ARM Cortex A54 CPU. To make an accurate and fair comparison, we use the following FPGA-implemented algorithms as baselines:
\begin{itemize}
    \item The \textit{QuantHD} algorithm from \cite{QuantHD}, on which \textit{QubitHD} is based
    \item Other state-of-the-art HD computing-based machine learning algorithms \cite{ISLPED, imani2019adapthd, morris2019comphd}
    \item A multi-level perceptron (MLP) \cite{sharma2016high} (see Table \ref{tab:compare})
    \item A binary neural network (BNN) \cite{umuroglu2017finn} (see Table \ref{tab:compare})
\end{itemize}
To show the advantage of the \textit{QubitHD} and the previous \cite{QuantHD} algorithm, we used the datasets summarized in Table \ref{tab:benchmark}. The datasets range from small datasets like UCIHAR and ISOLET (frequently used in IoT devices, for which \textit{QubitHD} is specially created), to larger datasets like face recognition.

\begin{table}[t!]
\centering
\caption{Datasets ($n$: number of features, $k$: number of classes).}
\label{tab:benchmark}
\resizebox{0.5\textwidth}{!}{
\begin{tabular}{c|cccccc}
\toprule
& $n$ & $K$ & \shortstack{\textbf{Data}\\ \textbf{Size}} & \shortstack{\textbf{Train}\\ \textbf{Size}} & \shortstack{\textbf{Test}\\ \textbf{Size}} & \textbf{Description}             \\ 
\hline
\textbf{ISOLET} & 617               & 26               & 19MB      & 6,238               & 1,559              & Voice Recognition~\cite{Isolet}              \\ 
\textbf{UCIHAR} & 561               & 12               & 10MB      & 6,213                & 1,554               & Activity recognition(Mobile)\cite{anguita2012human} \\ 
\textbf{MNIST} & 784                & 10                & 220MB     & 60,000             & 10,000            & Handwritten digits~\cite{lecun2010mnist}    \\ 
\textbf{FACE}   & 608  & 2  & 1.3GB     & 522,441  & 2,494  & Face recognition\cite{kim2017orchard}\\ 
\textbf{EXTRA}  & 225  & 4   & 140MB     & 146,869   & 16,343   & Phone position  recognition\cite{vaizman2017recognizing}\\ \hline
\end{tabular}
}
\end{table}

\subsection{Accuracy}\label{sec:accuracy}

\begin{table*}[t!]
\caption{Comparison of \textit{QubitHD} classification accuracies with the state-of-the-art HD computing.}
\centerline{
\resizebox{0.67\textwidth}{!}{
\begin{tabular}{ccc|cc|c}
\toprule
 & \multicolumn{2}{c|}{Baseline HD} & \multicolumn{2}{c|}{QuantHD} & \textit{QubitHD} \\
 & \textit{Non-Quantized} & \textit{Binary} & \textit{Non-Quantized} & \textit{Binary} & \textit{Binary with randomized flip} \\ \hline
\multicolumn{1}{l}{\textbf{ISOLET}} & 91.1\% & 88.1\% & 95.8\% & 94.6\% & 95.3\% \\
\textbf{UCIHAR} & 93.8\% & 77.4\% & 95.9\% & 93.0\% & 94.1\% \\
\textbf{MNIST} & 88.1\% & 32.70\% & 91.2\% & 87.1\% & 88.3\% \\
\textbf{FACE} & 95.9\% & 68.4\% & 96.2\% & 94.6\% & 95.4\% \\ \hline
\textbf{Mean} & \textbf{92.23\%} & \textbf{65.9\%} & \textbf{94.78\%} & \textbf{92.33\%} & \textbf{93.28} \\ \hline
\end{tabular}
}
}\label{tab:accuracy}
\end{table*}

The evaluation of the baseline HD model provides high-classification accuracy when using non-binarized hypervectors for classification. The problem, however, is that retraining and inference with a non-binary class hypervector is very costly, as it requires calculating cosine similarities. 
That is, for $k$-bit integers or floating-point numbers $\mathcal{O}(Dk^2)$ basic operations need to be performed through every step. These are costly and impractical on small-scale and battery-powered devices.

Similarly, during inference, the associative search between a query and a trained model requires the calculation of the costly cosine similarities. To address this issue, many HD computing-based machine learning algorithms binarize their models \cite{ISLPED}. That way, the cosine similarity is replaced by a simple \textit{Hamming distance} similarity check. The key problem with this approach is that it leads to a significant decrease in classification accuracy, as shown in Table \ref{tab:accuracy}.

The authors of \cite{QuantHD} already showed the existence of a partial solution to this problem, which involves simultaneously retraining the non-binarized model, while updating the binarized model. We already listed the problems with this model in subsection \ref{list:problems}. What especially motivated us to create a stable and more reliable \textit{QubitHD} algorithm, is the fact that the \textit{QuantHD} algorithm's retraining is unstable and unreliable. 
After extensively testing the \textit{QubitHD} algorithm, we conclude that it, on average, closes the gap of classification accuracy by 38.8\% as compared to the baseline HD computing-based machine learning algorithms in \cite{ISLPED} using the QuantHD framework (See Table~\ref{tab:accuracy}).

Additionally, we observe that the accuracies of the \textit{QubitHD} algorithm, using a binary model, are \textbf{1.2\%} and \textbf{60\%} higher than the classification accuracies of the baseline HD computing-based algorithm using non-quantized and binary respectively. Figure~\ref{fig:accuracyy} compares the classification accuracy of \textit{QubitHD} and QuantHD during different training iterations. It is clear that \textit{QubitHD} converges much faster than \textit{QuantHD}, as a result of the stochastic binarization process. 


\begin{table*}[htp]
\caption{Comparison of \textit{QubitHD} with MLP and BNN in terms of accuracy, efficiency, and model size \cite{QuantHD}}
\label{tab:compare}
\centerline{
\resizebox{0.98\textwidth}{!}{
\begin{tabular}{c|c|ccc|ccc|ccc|ccc}
\toprule
 & \textbf{\begin{tabular}[c]{@{}c@{}}MLP/BNN \\ Topologies\end{tabular}} & \multicolumn{3}{c|}{\textbf{Accuracy}}              & \multicolumn{3}{c|}{\textbf{CPU Training (s)}}   & \multicolumn{3}{c|}{\textbf{FPGA Inference ($\mu$s)}}    & \multicolumn{3}{c}{\textbf{Model Size}}           \\ 
                  &                                                                                         & \textit{MLP}    & \textit{BNN}    & \textit{\textit{QubitHD}}     & \textit{MLP}   & \textit{BNN}   & \textit{\textit{QubitHD}}    & \textit{MLP}     & \textit{BNN}    & \textit{\textit{QubitHD}}     & \textit{MLP}   & \textit{BNN}    & \textit{\textit{QubitHD}}     \\ \hline
\textbf{ISOLET}   & 617-512-256-26                                                                          & 95.8\%          & 96.1\%          & 95.3\%          & 2.08           & 17.69          & 0.19           & 27.39            & 5.24            & 0.28            & 1.81MB         & 56.7KB          & 65.0KB          \\ 
\textbf{UCIHAR}   & 561-512-256-12                                                                          & 97.3\%          & 95.9\%          & 94.1\%          & 1.04           & 8.32           & 0.08           & 21.43            & 5.18            & 0.27            & 1.68MB         & 52.7KB          & 30.0KB          \\ 
\textbf{FACE}     & 608-512-256-2                                                                           & 96.1\%          & 96.1\%          & 95.4\%          & 0.56           & 4.30           & 0.03           & 17.68            & 5.11            & 0.24            & 1.77MB         & 55.3KB          & 5.0KB          \\
\hline
\end{tabular}
}
}
\end{table*}

\subsection{Training Efficiency}
\begin{figure} [t!]
\centerline{
{\includegraphics[width=0.9\columnwidth]{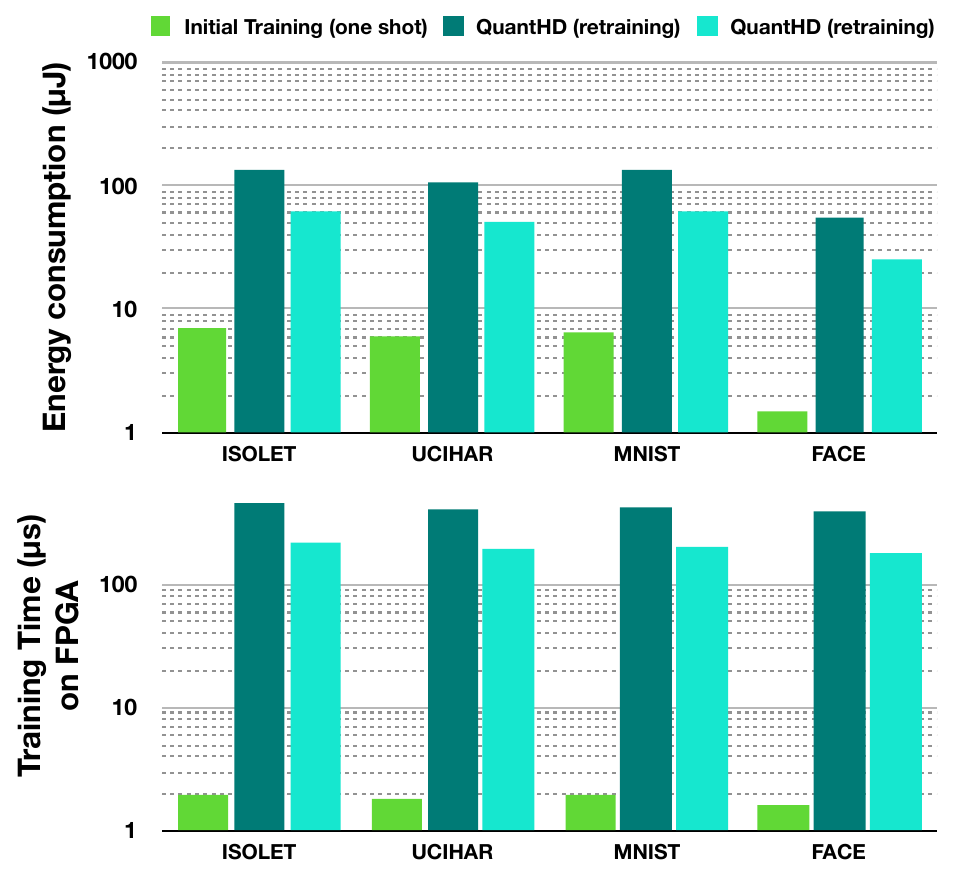}}
}
\caption{Energy consumption and execution time of initial training, \textit{QuantHD} retraining and \textit{QubitHD} retraining on a FPGA}
\label{fig:training}
\end{figure}

The training efficiency of HD-based algorithms is characterized by initial training of the model and subsequent retraining. Figure~\ref{fig:training} shows the energy consumption and execution time of \textit{QubitHD} during retraining. 
\begin{itemize}
    \item Algorithms in this type all consume the same energy during the generation of the initial training model
    \item The significant cost is the retraining: compared to the non-binarized model, QuantHD uses fewer operations when calculating the hypervector similarities (step 4 in Figure \ref{fig:framework}).
    \item No complex \textit{cosine} similarity has to be computed as calculating the Hamming distance is sufficient to determine whether there was a correct classification or not
    \item The improvement of \textit{QubitHD} lies in the faster convergence to a high classification accuracy, which is \textbf{30-50\%} faster than in \textit{QuantHD} and also decreases the energy consumption after the initial training proportionally.
    \item The \textit{QubitHD} modification has a dual benefit. It makes it possible to save energy and time during training, whilst achieving the same or better classification accuracies during testing depending on whether the goal is rapid convergence or high classification accuracy.
\end{itemize}

\subsection{Inference Efficiency}

Compared to QuantHD, there is no gain or loss in the time execution or energy efficiency, since the models behave identically once they are trained. So we report the same 44 $\times$ energy efficiency improvement and 5 $\times$ speedup as compared to the non-binarized HD algorithm.

\subsection{\textit{QubitHD} comparison with MLP and BNN}

\textit{QubitHD} is a classifier intended to run on low-powered devices, specifically with the goal of low energy consumption and fast and efficient execution in mind. As such, we set out to compare \textit{QubitHD}, not only to QuantHD but also to other non-HD lightweight classifiers. In our analysis, we compared \textit{QubitHD} accuracy and efficiency with the state-of-the-art lightweight classifiers, including Multi-Layer Perceptron (MLP) and Binarized Neural Network (BNN). 
For MLP and BNN, we aimed to use the same metric as employed in~\cite{umuroglu2017finn} with the small modification in input and output layers in order to run different applications. The results of this, presented in Table~\ref{tab:compare}, indicate that \textit{QubitHD}, while having similar classification accuracies to very lightweight classifier BNNs and MLPs, drastically reduces CPU usage during training and execution time during the inference. In particular, compared to MLPs \textit{QubitHD} uses 12 $\times$ less CPU during training and is 84 $\times$ faster during the inference on FPGAs. Compared with BNNs, \textit{QubitHD} uses a factor of 101 $\times$ less CPU time during training and is still 20 $\times$ faster during the inference.

\section{Conclusion}
Machine learning algorithms, based on Brain-inspired Hyperdimensional (HD) computing, imitate cognition by exploiting statistical properties of very high-dimensional vector spaces. They are a promising solution for energy-efficient classification tasks. A weakness of existing HD computing-based ML algorithms is the fact that they have to be binarized to achieve very high energy efficiency. At the same time, binarized models reach lower classification accuracies. In order to solve the problem of the trade-off between energy efficiency and classification accuracy, we propose the \textit{QubitHD} algorithm. With \textit{QubitHD}, it is possible to use binarized HD computing-based ML algorithms, which provide virtually the same classification accuracies as their non-binarized counterparts. The algorithm is inspired by stochastic quantum state measurement techniques. The improvement of \textit{QubitHD} is a duality and is reflected in the quicker convergence, and the higher and more stable classification accuracy achieved, as compared to \textit{QuantHD}.

\ifx{
\textbf{Our main contributions are}:
\begin{enumerate}
    \item The FPGA implementation of \textit{QubitHD} provides, on average, a 65\% improvement in terms of energy efficiency, and a 95\% improvement in terms of training time, as compared to the most recent state-of-the-art HD computing-based machine learning algorithm \textit{QuantHD} \cite{QuantHD}
    \item When compared with state-of-the-art low-cost classifiers like Binarized Neural Networks (BNN) and Multi-Layer Perceptrons (MLP), \textit{QubitHD} offers a similar classification accuracy whilst reducing training time by $56\times$ and allows for $52\times$ faster inference when testing. 
    \item \textit{QubitHD} decreases the classification accuracy gap between state-of-the-art binarized and non-binarized HD models by almost half. 
    \item \textit{QubitHD} converges, on average, $40\%$ faster during training, thus significantly decreasing the energy consumption.
\end{enumerate}
}\fi


\newpage

\bibliography{mybibliography}

\end{document}